\documentclass[11pt,a4paper]{article}
\usepackage[hyperref]{acl2017}
\usepackage{times}
\usepackage{latexsym}
\usepackage{url}
\usepackage{latexsym}
\usepackage{graphicx}
\usepackage{fancybox}
\usepackage{epsfig}
\usepackage{float}
\usepackage{amsmath,amsfonts,amssymb}
\usepackage{subfigure}
\usepackage{multirow}
\usepackage{bm}
\usepackage{color}
\usepackage{tikz-dependency}
\usepackage{xspace}

\aclfinalcopy % Uncomment this line for the final submission
\usepackage{amsmath,amsfonts,amssymb,bm}

\newcommand{\mat}[1]{\mathbf{#1}}
\newcommand{\trans}[1]{#1^{\textsf{T}}}

\usepackage{adjustbox}
\usepackage{array}
\usepackage{booktabs}

\newcolumntype{R}[2]{%
    >{\adjustbox{angle=#1,lap=\width-(#2)}\bgroup}%
    l%
    <{\egroup}%
}

\newcommand{\rela}[1]{{\sc #1}} % Discourse relation
\renewcommand{\trans}[1]{\ensuremath{#1^{\top}}}

\newcommand{\softmax}[1]{\text{softmax}\left(#1\right)}

\usepackage[normalem]{ulem}

\newcommand{\edurep}[1]{\mathbf{e}_{#1}}
\newcommand{\textrep}[1]{\mathbf{v}_{#1}}
\newcommand{\unlabeledm}{{\sc Unlabeled}\xspace}
\newcommand{\rootm}{{\sc Root}\xspace}
\newcommand{\fullm}{{\sc Full}\xspace}
\newcommand{\additionm}{{\sc Additive}\xspace}

\title{Neural Discourse Structure for Text Categorization}

\author{Yangfeng Ji {\rm and} Noah A.~Smith\\
  Paul G.~Allen School of Computer Science \& Engineering \\
  University of Washington \\
  Seattle, WA 98195, USA \\
  {\tt \{yangfeng,nasmith\}@cs.washington.edu} \\
}

\date{}

\begin{document}
\maketitle
\begin{abstract}
We show that discourse structure, as defined by Rhetorical Structure Theory and provided by an existing discourse parser, benefits text categorization.  Our approach uses a recursive neural network and a newly proposed attention mechanism to compute a representation of the text that focuses on salient content, from the perspective of both RST and the task.  Experiments consider variants of the approach and illustrate its strengths and weaknesses.

% We introduce discourse structure derived from the Rhetorical Structure Theory as an inductive bias, since by nature it reflects the organization of a text. 
% Then, a recursive neural network model with a newly proposed attention mechanism is constructed based on the discourse structure, and used to composite the distributed representation of this text. 
% Experiments on four text categorization tasks show our models outperform prior work on most of the corpora.
% Additional analysis exhibits the limitation of this work on other genres of texts, and also indicates the benefit of getting a better discourse parser.

\end{abstract}

% *****************************************************************
\section{Introduction}
\label{sec:intro}

Advances in text categorization have the potential to improve systems for analyzing sentiment, inferring authorship or author attributes, making predictions, and many more.  
% While a problem: limited amounts of annotated data for training, but a certain amount of predictability in the structure of a document in the genre of interest.
Several past researchers have noticed that methods that reason about the relative salience or importance of passages within a text can lead to improvements \citep{ko2004improving}.  Latent variables
\citep{yessenalina2010multi}, structured-sparse regularizers \citep{yogatama2014linguistic}, and neural attention models \citep{yang2016hierarchical} have all been explored.

\begin{figure}
  \centering
  \includegraphics[width=0.45\textwidth]{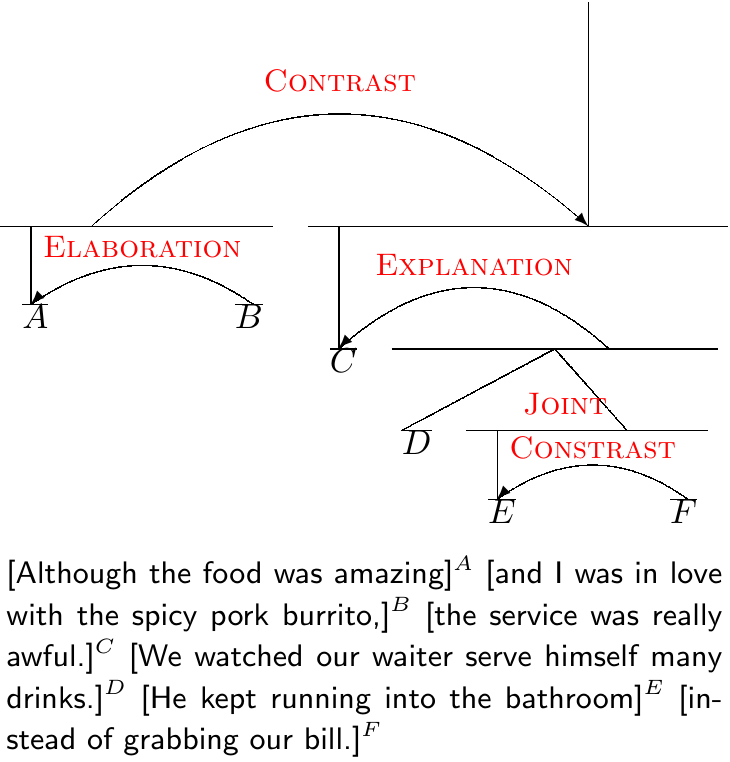}
  \caption{A manually constructed example of the RST~\citep{mann1988rhetorical} discourse structure on a text.}
  \label{fig:rst-example}
\end{figure}

\textbf{Discourse structure}, which represents the organization of a text as a tree (for an example, see \autoref{fig:rst-example}), 
might provide cues for the importance of different parts of a text.
Some promising results on sentiment classification tasks support this idea:
\citet{bhatia2015better} and \citet{hogenboom2015using} applied hand-crafted weighting schemes to the sentences in a document, based on its discourse structure, and showed benefit to sentiment polarity classification.

In this paper, we investigate the value of discourse structure for text categorization more broadly, considering five tasks, through the use of a recursive neural network built on an automatically-derived document parse from a top-performing, open-source discourse parser, DPLP \citep{ji2014representation}.  
Our models learn to weight the importance of a document's sentences, based on their positions and relations in the discourse tree.  
We introduce a new, unnormalized attention mechanism to this end.

Experimental results show that variants of our model outperform prior work on four out of five tasks considered.   Our method unsurprisingly underperforms on the fifth task, making predictions about legislative bills---a genre in which discourse conventions are quite different from those in the discourse parser's training data.
Further experiments show the effect of discourse parse quality on text categorization performance, suggesting that future improvements to discourse parsing will pay off for text categorization, and validate our new attention mechanism. 

Our implementation is available at \url{https://github.com/jiyfeng/disco4textcat}.

% A few more references to show that the structural sparsity has a long history:
% \begin{itemize}
% \item \citep{pang2004sentimental}: find the subjective part of a review for sentiment polarity prediction
% \item \citep{lin1997identifying}: identify the topic-bearing sentence from a text with position information
% \end{itemize}

% ------------------------------
\section{Background: Rhetorical Structure Theory}
\label{sec:discourse}

Rhetorical Structure Theory \citep[RST;][]{mann1988rhetorical} is a theory of discourse
that has enjoyed popularity in NLP.   RST posits that a document can be represented by a tree whose leaves are elementary discourse units (EDUs, typically clauses or sentences).  Internal nodes in the tree correspond to spans of sentences that are connected via discourse relations such as \rela{Contrast} and \rela{Elaboration}.  In most cases, a discourse relation links adjacent spans denoted ``nucleus'' and ``satellite,'' with the former more essential to the writer's purpose than the latter.\footnote{There are also a few exceptions in which a relation can be realized with multiple nuclei.}

An example of a manually constructed RST parse 
for a restaurant review is shown in \autoref{fig:rst-example}.  The six EDUs are indexed from $A$ to $F$; the discourse tree organizes them hierarchically into increasingly larger spans, with the last \rela{Contrast} relation resulting in a span that covers the whole review.
Within each relation, the RST tree indicates the nucleus pointed by an arrow from its satellite (e.g., in the \rela{Elaboration} relation, $A$ is the nucleus and $B$ is the satellite).

The information embedded in RST trees has motivated many applications in NLP research, including document summarization~\citep{marcu1999discourse}, argumentation mining~\citep{azar1999argumentative}, and sentiment analysis~\citep{bhatia2015better}.
In most applications, RST trees are built by automatic discourse parsing, due to the expensive cost of manual annotation.  In this work, we use a state-of-the-art open-source RST-style discourse parser, DPLP \cite{ji2014representation}.\footnote{\url{https://github.com/jiyfeng/DPLP}}

\begin{figure*}
  \centering
  \subfigure[dependency structure]{
    \includegraphics[width=0.30\textwidth]{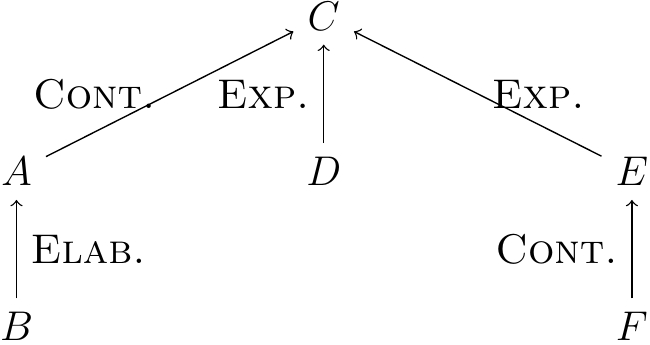}
    \label{fig:dep-example}
  }
  \subfigure[recursive neural network structure]{
    \includegraphics[width=0.60\textwidth]{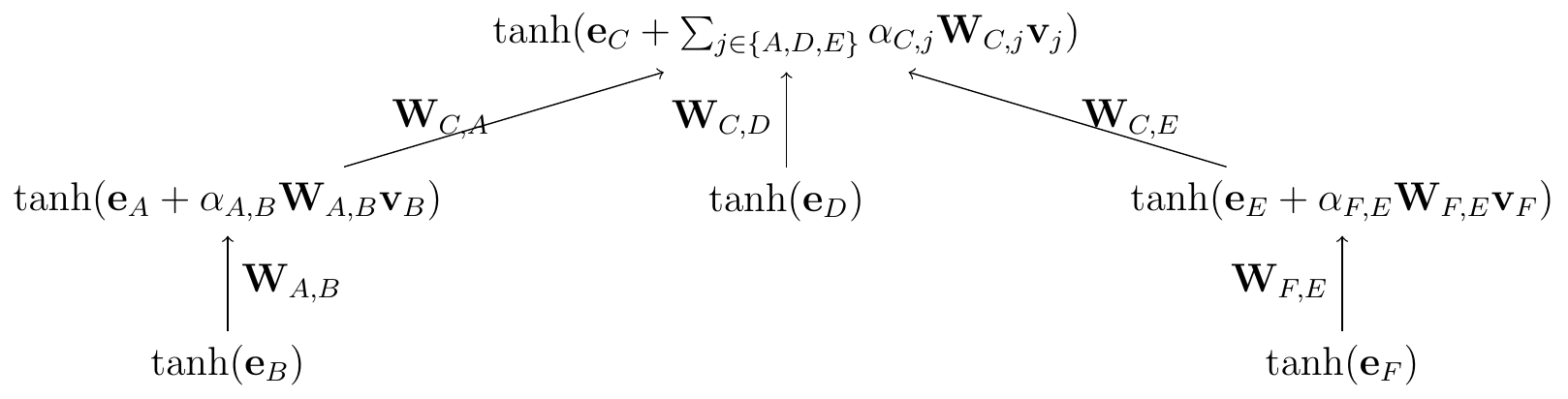}
    \label{fig:dep-network}
  }
  \caption{The dependency discourse tree derived from the example RST
    tree in \autoref{fig:rst-example} (a) and the corresponding
    recursive neural network model on the tree (b).
  }
  \label{fig:dep}
\end{figure*}

We follow recent work that suggests transforming the RST tree into a dependency  structure~\citep{yoshida2014dependency}.\footnote{The transformation is trivial and deterministic given the nucleus-satellite mapping for each relation.  The procedure is analogous to the transformation of a headed phrase-structure parse in syntax into a dependency tree \citep[e.g.,][]{yamada-03}.}
Figure~\autoref{fig:dep-example} shows the corresponding dependency structure of the RST tree in \autoref{fig:rst-example}.  It is clear that $C$ is the root of the tree, and in fact this clause summarizes the review and suffices to categorize it as negative.
This dependency representation of the RST tree offers a form of inductive bias for our neural model, helping it to discern the most salient parts of a text in order to assign it a label.

% ------------------------------
\section{Model}
\label{sec:model}

Our model is a recursive neural network built on a discourse
dependency tree.
It includes a distributed representation computed for each EDU, and a
composition function that combines EDUs and partial trees into larger
trees.  At the top of the tree, the representation of the complete
document is used to make a categorization decision.
Our approach is analogous to (and inspired by) the use of recursive neural
networks on \emph{syntactic} dependency trees, with word embeddings at
the leaves \citep{socher2014grounded}.

\subsection{Representation of Sentences}
\label{subsec:edu}

Let $\edurep{}$ be the distributed representation of an EDU.
We use a bidirectional LSTM on the words' embeddings within each EDU (details of word
embeddings are given in \autoref{sec:imp}), concatenating the last hidden state vector from the
forward LSTM ($\overrightarrow{\edurep{}}$) with that of the backward
LSTM ($\overleftarrow{\edurep{}}$) to get~$\edurep{}$.

There is extensive recent work on architectures for embedding
representations of sentences and other short pieces of text,
including, for example, 
(bi)recursive neural networks~\citep{paulus2014global} and
convolutional neural networks~\citep{kalchbrenner2014convolutional}.
Future work might consider alternatives; we chose the
bidirectional LSTM due to its effectiveness in many settings.
%\yfcomment{Do we need citations here?}

\subsection{Full Recursive Model}

Given the discourse dependency tree for an input text, our recursive
model builds a vector representation through composition at each arc
in the tree.  Let $\textrep{i}$ denote the vector representation of
EDU $i$ and its descendants. 
For the base case where EDU $i$ is a leaf in the tree, we let
$\textrep{i}=\tanh(\edurep{i})$, which is the elementwise hyperbolic tangent function.

For an internal node $i$, the composition function considers a
parent and all of its children, whose indices are denoted by $\mathit{children}(i)$.
In defining this composition function, we seek for (i.)  the contribution of the
parent node $\edurep{i}$ to be central; and (ii.) the contribution of
each child node $\edurep{j}$ be determined by its content as well as
the discourse relation it holds with the parent.  We therefore define
\begin{equation}
  \textrep{i} = \tanh \left
    (\edurep{i} +\sum_{j \in \mathit{children}(i)} \alpha_{i,j}\mathbf{W}_{r_{i,j}}\textrep{j}
    \right),   
    \label{eq:comp-func}
\end{equation}
where $\mathbf{W}_{r_{i,j}}$ is a relation-specific composition matrix
indexed by the relation between $i$ and $j$, $r_{i,j}$. 
% \nascomment{important:  what is $\bm{f}$?  I think that
%   ``$\bm{f}(\edurep{i})$'' should actually be ``$\edurep{i}$''.}
% \nascomment{should the above equation use $\textrep{j}$ for the
%   children instead of $\edurep{j}$?}

$\alpha_{i,j}$ is an ``attention'' weight, defined as
\begin{equation}
  \label{eq:attention}
  \alpha_{i,j} = \sigma
  \left(\trans{\edurep{i}}\mat{W}_{\alpha}\textrep{j} \right),
\end{equation}
% \nascomment{why $\bm{f}$?  and why is it transposed??  shouldn't it
%   just be $\textrep{i}^\top$ like it is in the other attention version
%   later in the paper?}
where $\sigma$ is the elementwise sigmoid and $\mat{W}_{\alpha}$
contains attention parameters (these are  relation-independent).
Our attention mechanism differs from prior work
\citep{bahdanau2015neural}, in which attention weights are normalized
to sum to one across competing candidates for attention.  Here,
$\alpha_{i,j}$ does not depend on node $i$'s other children.  This is
motivated by RST, in which the presence of a node does not signify
lesser importance to its siblings.  Consider, for example, EDU $D$
and text span $E\text{-}F$ in \autoref{fig:rst-example}, which in parallel provide
\rela{Explanation} for EDU $C$.  This scenario differs from machine
translation,  where attention isused to implicitly and softly align
output-language words to relatively few input-language words.  It also
differs from attention in composition functions used in syntactic
parsing \citep{kuncoro-17}, where attention can mimic head rules that follow from an
endocentricity hypothesis of syntactic phrase representation.  

Our recursive composition function, through the attention mechanism
and the relation-specific weight matrices, is designed to learn how to
differently weight EDUs for the categorization task.
This idea of using a weighting scheme along with discourse structure
is explored in prior works
\citep{bhatia2015better,hogenboom2015using}, although they are
manually designed, rather than learned from training data.

Once we have $\textrep{\textit{root}}$ of a text, the prediction of
its category is given by 
$\softmax{\mat{W}_o\textrep{\textit{root}}+\mathbf{b}}$.

We refer to this model as the \fullm model, since it makes use of the entire
discourse dependency tree.

\subsection{Unlabeled Model}
The \fullm model based on \autoref{eq:comp-func} uses a dependency discourse
tree with relations.  
Because alternate discourse relation labels have been
proposed \citep[e.g.,][]{prasad2008penn}, we seek to measure the effect of these
labels. 
% \yfcomment{not sure I get this line. PDTB does provide a alternative way
% to categorize discourse relations. 
% But it is not clear to me how this connects to the unlabeled model?}
% \nascomment{the point is that the labels are not set in stone; people
%   disagree about what the labels should be.  maybe we should just let
%   the model figure out implicit labels?}
We therefore consider an \unlabeledm model based only on the
tree structure, without the relations:
\begin{equation}
  \label{eq:comp-func-2}
  \textrep{i} = \tanh\left(\edurep{i}+\sum_{j \in
      \mathit{children}(i)}\alpha_{i,j}\textrep{j} \right).
\end{equation}
% \nascomment{changed the above equation, which used $\textrep{i}$ on
%   the right (making it circular!).  I replaced $\textrep{i}$ with $\edurep{i}$.}
Here, only attention weights are used to compose the children nodes'
representations, significantly reducing the number of model parameters.

This \unlabeledm model is similar to the depth weighting scheme
introduced by \citet{bhatia2015better}, which also uses an unlabeled
discourse dependency tree, but our attention weights are computed by a
function whose parameters are learned.  This approach sits squarely
between \citet{bhatia2015better} and the flat document structure used
by \citet{yang2016hierarchical}; the \unlabeledm model still uses discourse to
bias the model toward some content (that which is closer to the tree's root).

\subsection{Simpler Variants}
We consider two additional baselines that are even simpler.  The
first, \rootm,
uses the discourse dependency structure only to select the root EDU,
which is used to represent the entire text:
% \begin{equation}
%   \label{eq:model-variant-2}
%   \textrep{\textit{root}}=\edurep{\textit{root}}. 
% \end{equation}
$ \textrep{\textit{root}}=\edurep{\textit{root}}.$
No composition function is needed.
This model variant is motivated by work on document summarization~\citep{yoshida2014dependency}, 
where the most central EDU is used to represent the whole text.

The second variant, \additionm, uses all the EDUs with a simple
composition function, and does not depend on discourse structure at
all:
% \begin{equation}
%   \label{eq:addition-model}
%   \textrep{\textit{root}} = \frac{1}{N}\sum_{i=1}^{N}\edurep{i},
% \end{equation}
$\textrep{\textit{root}} = \frac{1}{N}\sum_{i=1}^{N}\edurep{i},$
where $N$ is the total number of EDUs.  This serves as a baseline to
test the benefits of discourse, controlling for other design decisions
and implementation choices.  Although sentence representations $\edurep{i}$ are 
built in a different way from the work of~\citet{yang2016hierarchical},
this model is quite similar to their HN-AVE model on building document representations.

% ------------------------------
\section{Implementation Details}
\label{sec:imp}
The parameters of all components of our model (top-level classification, composition, and EDU representation) are learned end-to-end using standard methods.
We implement our learning procedure with the {\tt DyNet} package~\citep{neubig2017dynet}. 

\paragraph{Preprocessing.}  For all datasets, we use the same 
preprocessing steps, mostly following recent work on language modeling
\citep[e.g.,][]{mikolov2010recurrent}.
We  lowercased all the tokens and removed tokens that contain only punctuation symbols.
We replaced numbers in the documents with a special number token.  
Low-frequency word types were replaced by \textsc{unk}; we reduce the vocabulary for each dataset until 
approximately 5\% of tokens are mapped to \textsc{unk}.
The vocabulary sizes after preprocessing are also shown in~\autoref{tab:datasets}.

\paragraph{Discourse parsing.} 
Our model requires the discourse structure for each document.
We used DPLP, the RST parser from~\citet{ji2014representation},
which is one of the best discourse parsers on the RST discourse treebank benchmark~\citep{carlson2001building}. 
It employs a greedy decoding algorithm for parsing, producing 2,000 parses per minute on average on a single CPU.
DPLP provides discourse segmentation, breaking a text into EDUs, typically clauses or sentences, based on syntactic parses provided by Stanford CoreNLP.
RST trees are converted to dependencies following the method of \citet{yoshida2014dependency}.    
DPLP as distributed is trained on 347 \emph{Wall Street Journal} articles from the Penn Treebank~\citep{marcus1993building}.

\paragraph{Word embeddings.}  In cases where there are 10,000 or fewer training examples,
we used pretrained GloVe word embeddings \citep{pennington2014glove}, 
following previous work on neural discourse processing \citep{ji2015one}.  
For larger datasets, we randomly initialized word embeddings and trained them alongside other model parameters.

\paragraph{Learning and hyperparameters.} 
Online learning was performed with the optimization method and initial learning rate as hyperparameters. 
To avoid the exploding gradient problem, we used the norm clipping trick with a threshold of $\tau=5.0$. 
In addition, dropout rate 0.3 was used on both input and hidden layers to avoid overfitting.
We performed grid search over the word vector representation dimensionality, the LSTM hidden state dimensionality (both $\{32, 48, 64, 128, 256\}$), the initial learning rate ($\{0.1, 0.01, 0.001\}$), and the update method (SGD and Adam, \citealp{kingma2014adam}).
For each corpus, the highest-accuracy combination of these hyperparameters is selected using development data or ten-fold cross validation, which will be specified in~\autoref{sec:datasets}.

% ------------------------------
\section{Datasets}
\label{sec:datasets}

\begin{table*}
  \centering
  {% \small
    \begin{tabular}{p{0.15\textwidth}crrrrrr}
      \toprule
      & & & \multicolumn{4}{c}{Number of docs.} & \\
      \cmidrule(l){4-7}
      Dataset & Task & Classes & Total & Training & Development & Test & Vocab.~size\\
      \midrule
      Yelp & Sentiment & 5 & 700K & 650K & -- & 50K & 10K \\
      MFC & Frames & 15 & 4.2K & -- & -- & --  & 7.5K \\
      Debates & Vote & 2 & 1.6K & 1,135 & 105 & 403  & 5K \\
      Movies & Sentiment & 2 & 2.0K & -- & -- & --  & 5K \\
      Bills & Survival & 2 & 52K & 46K & -- & 6K & 10K\\
      \bottomrule
    \end{tabular}
  }
  \caption{Information about the five datasets used in our
    experiments. To compare with prior work, we use different experimental settings.
    For Yelp and Bill corpora, we use 10\% of the training examples 
    as development data. For MFC and Movies corpora, we use 10-fold cross validation and report 
    averages across all folds.}
  \label{tab:datasets}
\end{table*}

% Checklist
% \begin{itemize}
% \item data description
% \item data split
% \item compared method --- what they did
% \end{itemize}

We selected five datasets of different sizes and corresponding to varying categorization tasks.
Some information about these datasets is summarized in~\autoref{tab:datasets}.

\paragraph{Sentiment analysis on Yelp reviews.} Originally from the Yelp Dataset Challenge in 2015,
this dataset contains 1.5 million examples.
We used  the preprocessed dataset from \citet{zhang2015character}, which has 650,000 training and 50,000 test examples. 
The task is to predict an ordinal rating (1--5) from the text of the review.
To select the best combination of hyperparameters, we randomly sampled 10\% training examples as the development data. 
We compared with hierarchical attention networks~\citep{yang2016hierarchical}, which use the normalized attention mechanism on both word and sentence layers with a flat document structure, and provide the state-of-the-art result on this corpus.

\paragraph{Framing dimensions in news articles.}
The Media Frames Corpus \citep[MFC;][]{card2015media} includes around 4,200 news articles about immigration from 13 U.S.~newspapers over the years 1980--2012. 
The annotations of these articles are in terms of a set of 15 general-purpose labels, such as \textsc{Economics} and \textsc{Morality}, designed to categorize the emphasis framing applied to the immigration issue within the articles.
We focused on predicting the single \emph{primary} frame of each article.
The state-of-the-art result on this corpus is from \citet{card2016analyzing}, where they used logistic regression together with unigrams, bigrams and Bamman-style personas \citep{bamman2014learning} as features. 
The best feature combination in their model alongside other hyperparameters was identified by a Bayesian optimization method~\citep{bergstra2015hyperopt}.
To select hyperparameters, we used a small set of examples from the corpus as a development  set.
Then, we report average accuracy across 10-fold cross validation as in~\citep{card2016analyzing}.

\paragraph{Congressional floor debates.} The corpus was originally collected by~\citet{thomas2006get},
and the data split we used was constructed by~\citet{yessenalina2010multi}.
The goal is to predict the vote (``yea'' or ``nay'') for the speaker of each speech segment.
The most recent work on this corpus is from~\citet{yogatama2014linguistic}, which proposed structured regularization methods based on linguistic components, e.g., sentences, topics, and syntactic parses.
Each regularization method induces a linguistic bias to improve text classification accuracy, where the best result we repeated here is from the model with sentence regularizers.

\paragraph{Movie reviews.} This classic movie review corpus was constructed by~\citet{pang2004sentimental} and includes 1,000 positive and 1,000 negative reviews.
On this corpus, we used the standard ten-fold data split for cross validation and reported the average accuracy across folds.
We compared with the work from both~\citet{bhatia2015better} and \citet{hogenboom2015using}, which are two recent works on discourse for sentiment analysis.
\citet{bhatia2015better} used a hand-crafted weighting scheme to bias the bag-of-word representations on sentences.
\citet{hogenboom2015using} also considered manually-designed weighting schemes and a lexicon-based model as classifier, achieving performance inferior to fully-supervised methods like \citet{bhatia2015better} and ours.
% In addition, the varying experimental setups on this corpus prevent direct comparison with some previous works~\citep[e.g.,][]{zaidan2007using} 
% \nascomment{this will raise eyebrows; too vague, and you say it as if we could not have tried comparing with other papers, when we could have replicated those setups.  what's the real issue here?}.
% \yfcomment{My word is confusing and sounds like an excuse. The real issue here is that this paper and some other papers using this dataset are not closely relevant. Another reason is that they may use a differrent data split, which is not the major issue. We can use the same split if we have to compare with them.}

\paragraph{Congressional bill corpus.} This corpus, collected by~\citet{yano2012textual}, includes 51,762 legislative bills from the 103rd to 111th U.S.~Congresses. 
The task is to predict whether a bill will survive based on its content.  
We randomly sampled 10\% training examples as development data to search for the best hyperparameters. 
To our knowledge, the best published results are due to \citet{yogatama2014linguistic}, which is the same baseline as for the congressional floor debates corpus.

% ------------------------------
\section{Experiments}
\label{sec:exp}

\begin{table*}
  \centering
  {% \small
  \begin{tabular}{p{0.35\textwidth}p{0.08\textwidth}p{0.08\textwidth}p{0.08\textwidth}p{0.08\textwidth}p{0.08\textwidth}}
    \toprule
    Method & Yelp & MFC & Debates & Movies & Bills \\
    \midrule
    \emph{Prior work}\\
    1. \citet{yang2016hierarchical} & 71.0 & --- & --- & --- & --- \\ 
    2. \citet{card2016analyzing} & --- & 56.8 & --- & --- & ---\\ 
    3. \citet{yogatama2014linguistic} & --- & --- & 74.0 & ---  & 88.5 \\
    4. \citet{bhatia2015better} & ---  & ---  & --- & 82.9 & --- \\
    5. \citet{hogenboom2015using} & --- & --- & --- & 71.9 & --- \\[0.2em]
    \emph{Variants of our model}\\
    6. \additionm  & 68.5 & \bf 57.6 & 69.0 & 82.7 & 80.1\\
    7. \rootm & 54.3 & 51.2 & 60.3 & 68.7  & 70.5 \\
    8. \unlabeledm  & \bf 71.3 & {\bf 58.4} & {\bf 75.7} & {\bf 83.1}  & 78.4 \\
    9. \fullm & {\bf 71.8} & 56.3 & \bf 74.2 & 79.5  & 77.0 \\
    \bottomrule
  \end{tabular}
}
  \caption{Test-set accuracy across five datasets.  Results from prior
    work are reprinted from the corresponding publications. Boldface
    marks performance stronger than the previous state of the art. 
    \label{tab:results}}
\end{table*}

We evaluated all variants of our model on the five datasets presented in~\autoref{sec:datasets}, comparing in each case to the published state of the art as well as the most relevant works.
% \yfcomment{For the five datasets, I was only able to run the binomial test on three of them --- for Movies and MFC, I was not sure how to do the test because of the cross validation. Among the rest three, only on the yelp corpus, the \fullm model against prior work is significant ($p < 0.05$).} \nascomment{ok ...}

\paragraph{Results.}  
See \autoref{tab:results}. 
% \yfcomment{Not sure the following two-word sentence is appropriate :)}
On four out of five datasets, our \unlabeledm model (line 8) outperforms past methods.  
In the case of the very large Yelp dataset, our \fullm model (line 9) gives even stronger performance, but not elsewhere, suggesting that it is overparameterized for the smaller datasets.  
Indeed, on the MFC and Movies tasks, the discourse-ignorant \additionm outperforms the \fullm model.
% \nascomment{should make this more concrete.  how many parameters for the full and unlabeled models, in the selected model for each dataset?}
% \yfcomment{I am not sure how to this. I can either use a table to show all the selected models, or give an equation in model section with respect to the dimensions and vocab size.} 
% \nascomment{I don't know if we need to give the model size systematically for every case.  here, it would be nice to say, for MFC and Movies, something like ``on these datasets, the selected full model had ten times as many parameters as the unlabeled model, which in turn had five times as many parameters as the additive model.''}
On these datasets, the selected \fullm model had nearly 20 times as many parameters as the \unlabeledm model, which in turn had twice as many parameters as the \additionm.
% \yfcomment{Another difference between unlabeled and additive is the model architecture, which I think is more important than the number of parameters, but I am not sure how to say this.}

This finding demonstrates the benefit of explicit discourse structure---even the output from an imperfect parser---for text categorization in some genres.
  This benefit is supported by both \unlabeledm and \fullm, since both of them use discourse structures of texts.
  The advantage of using discourse information varies on different genres and different corpus sizes.
Even though the discourse parser is trained on news text, it still offers benefit to restaurant and movie reviews and to the genre of congressional debates.   
Even for news text, if the training dataset is small (e.g., MFC), a lighter-weight variant of discourse (\unlabeledm) is preferred.

Legislative bills, which have technical legal content and highly specialized conventions (see the supplementary material for an example), are arguably the most distant genre from news among those we considered.
On that task,  we see discourse working against accuracy.  
Note that the corpus of bills is more than ten times larger than three cases where our \unlabeledm model outperformed past methods, suggesting that the drop in performance is not due to lack of data.

It is also important to notice that the \rootm model performs quite poorly in all cases.  
This implies that discourse structure is not simply helping by finding a single EDU upon which to make the categorization decision.

\begin{figure*}
  \centering
  \subfigure[true label: 2, predicted label: 2]{
    \includegraphics[width=0.45\textwidth]{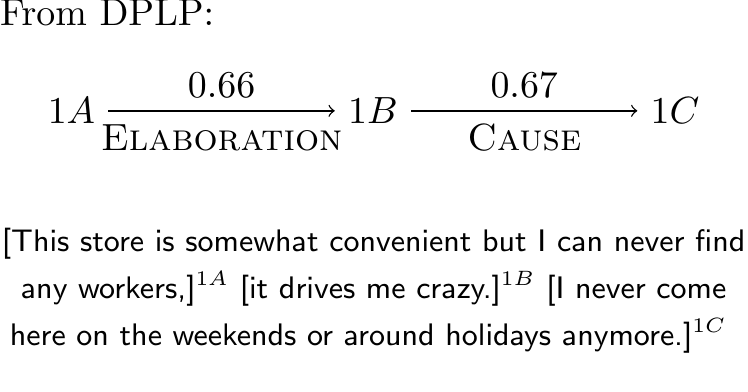}
    \label{fig:yelp-ex-1}
  }\\
  \subfigure[true label: 5, predicted label: 5]{
    \includegraphics[width=0.6\textwidth]{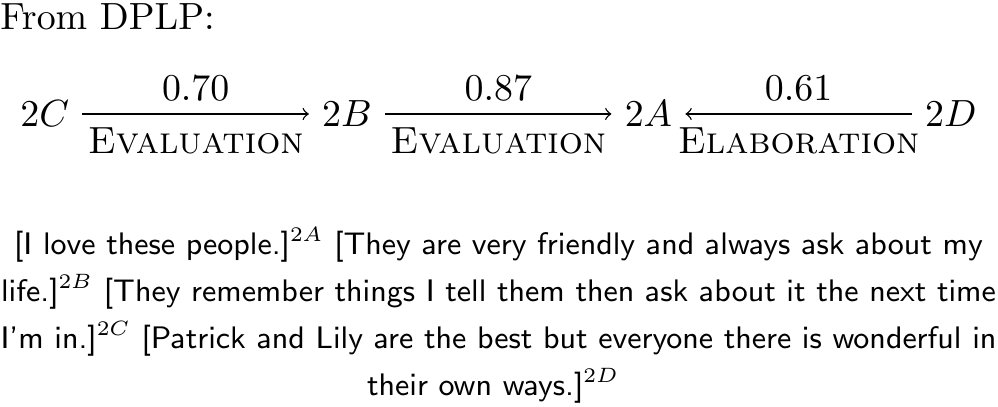}
    \label{fig:yelp-ex-2}
  }
  \par\bigskip
  \subfigure[true label: 1, predicted label: 3]{
    \includegraphics[width=0.9\textwidth]{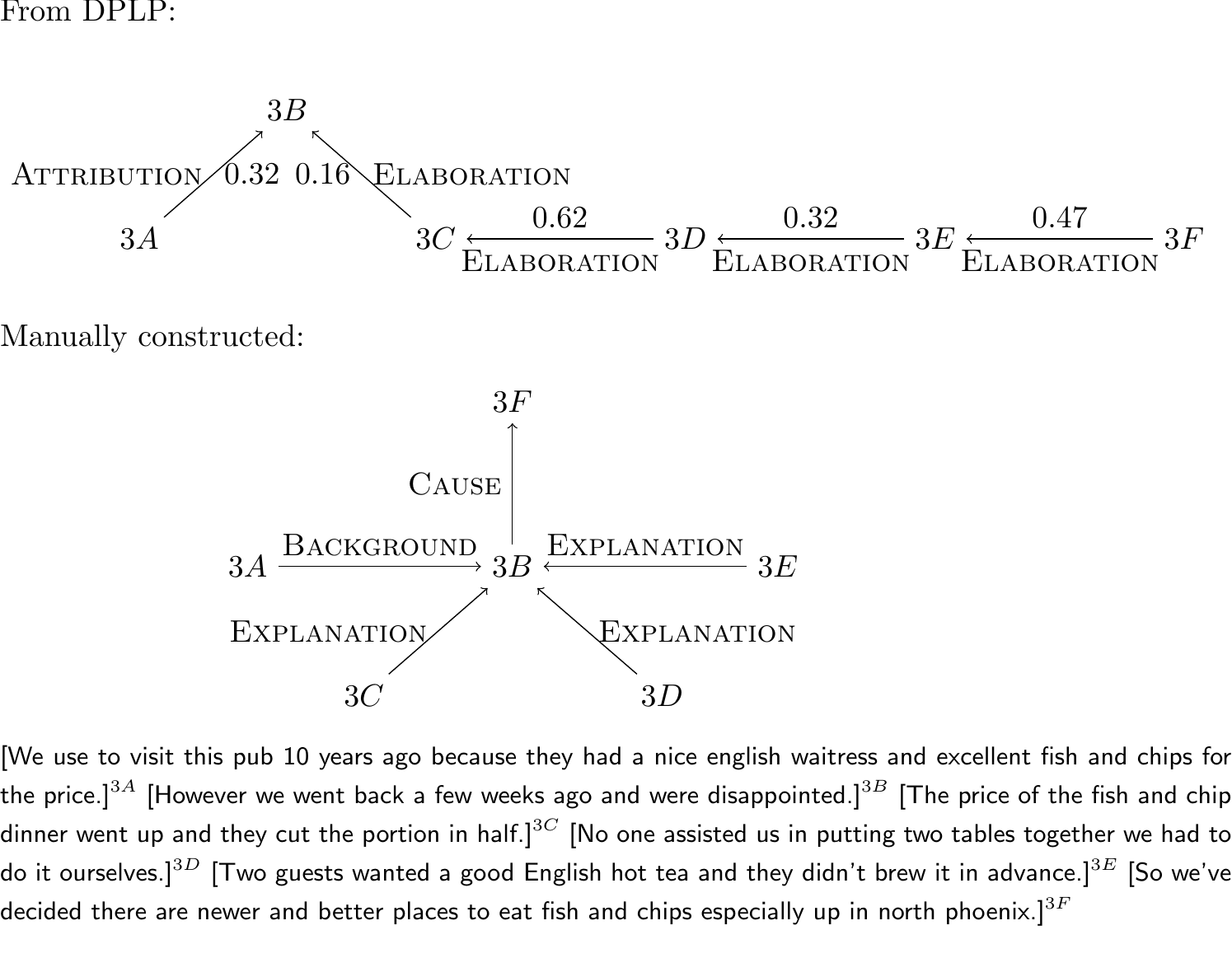}
    \label{fig:yelp-ex-3}
  }
  \caption{Some example texts (with light revision for readability)
    from the Yelp Review corpus and their corresponding dependency
    discourse parses from DPLP~\citep{ji2014representation}. 
    The numbers on dependency edges are attention weights 
    produced by the \fullm model. 
    % \nascomment{I don't understand what the numbers
    %   are on the trees.  could we put the relation labels in, somehow?}
    % \yfcomment{The numbers are attention weights. I didn't add the discourse relations here, 
    %   because these numbers are relation independent. Showing them together may lead to a confusion.
    %   What do you think?}
  \label{fig:yelp-examples}}
\end{figure*}

\paragraph{Qualitative analysis.} 
% \yfcomment{I am not confident with this paragraph, mainly because don't know what I should say here.}
\autoref{fig:yelp-examples} shows some example texts from the Yelp Review corpus with their discourse structures produced by DPLP, where the weights were generated with the \fullm model.
Figures \autoref{fig:yelp-ex-1} and \autoref{fig:yelp-ex-2} are two successful examples of the \fullm model.
Figure~\autoref{fig:yelp-ex-1} shows a simple case with respect to the discourse structure.
Figure~\autoref{fig:yelp-ex-2} is slightly different---the text in this example may have more than one reasonable discourse structure, e.g., $2D$ could be a child of $2C$ instead of $2A$.
In both cases, discourse structures help the \fullm model bias to the important sentences.

Figure~\ref{fig:yelp-ex-3}, on the other hand, presents a negative example, where DPLP failed to identify the most salient sentence $3F$.
In addition, the weights produced by the \fullm model do not make much sense, which we suspect the model was confused by the structure. 
Figure~\autoref{fig:yelp-ex-3} also presents a manually-constructed discourse structure on the same text for reference. 
A more accurate prediction is expected if we use this manually-constructed discourse structure, because it has the appropriate dependency between sentences.
In addition, the annotated discourse relations are able to select the right relation-specific composition matrices in \fullm model, which are consistent with the training examples.
% \nascomment{what happens if you feed *this* parse into the classifier?}

\begin{figure}
  \centering
  \includegraphics[width=0.45\textwidth]{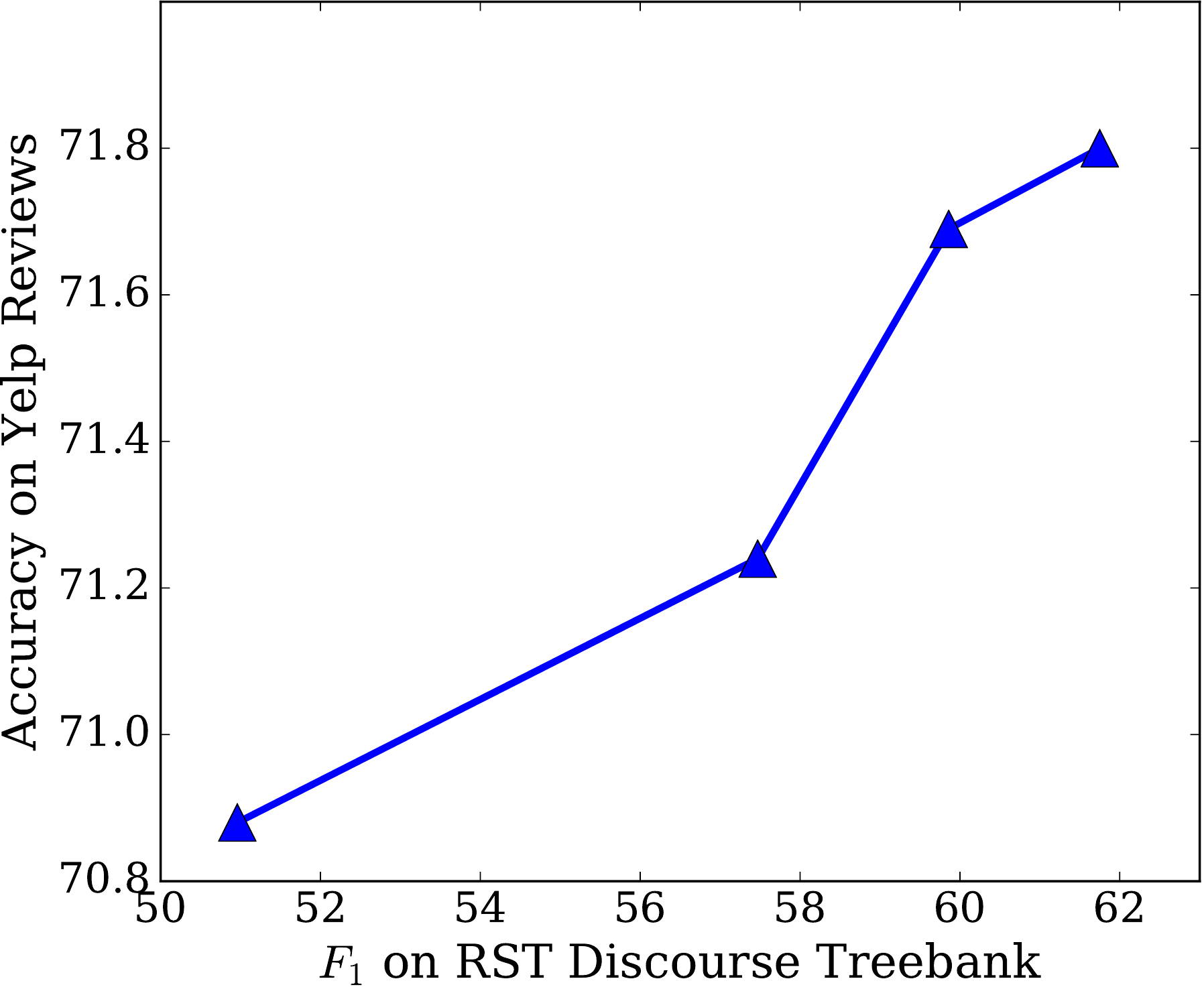}
  \caption{Varying the amount of training data for the discourse parser, we can see how parsing $F_1$ performance affects accuracy on the Yelp review task.}
  \label{fig:dis-accuracy}
\end{figure}

\paragraph{Effect of parsing performance.}
A natural question is whether further improvements to RST discourse parsing would lead to even greater gains in text categorization.  
While advances in discourse parsing are beyond the scope of this paper, we can gain some insight by exploring degradation to the DPLP parser.  
An easy way to do this is to train it on subsets of the RST discourse treebank. 
We repeated the conditions described above for our \fullm model, training DPLP on 25\%, 50\%, and 75\% of the training set (randomly selected in each case) before re-parsing the data for the sentiment analysis task.  
We did not repeat the hyperparameter search.
In~\autoref{fig:dis-accuracy},  we plot accuracy of the classifier ($y$-axis) against the $F_1$ performance of the discourse parser ($x$-axis).  
Unsurprisingly, lower parsing performance implies lower classification accuracy.  
Notably, if the RST discourse treebank were reduced to 25\% of its size, our method would underperform the discourse-ignorant model of \citet{yang2016hierarchical}.  
While we cannot extrapolate with certainty, these findings suggest that further improvements to discourse parsing, through larger annotated datasets or improved models, could lead to greater gains.

\paragraph{Attention mechanism.}
In~\autoref{sec:model}, we contrasted our new attention mechanism (\autoref{eq:attention}), which is inspired by RST's lack of ``competition'' for salience among satellites, 
with the attention mechanism used in machine translation \citep{bahdanau2015neural}.  
We consider here a variant of our model with normalized attention:
\begin{equation}
  \label{eq:alternative-attention}
  \bm{\alpha}'_{i} =\text{softmax}\left( \left[ 
      \begin{array}{c} 
        \vdots\\ 
        \trans{\textrep{j}} \\ 
        \vdots 
      \end{array} \right ]_{j \in \mathit{children}(i)} 
    \mat{W}_{\alpha}\cdot\edurep{i} \right) .
\end{equation}
% \nascomment{this isn't consistent with the earlier attention.  are we mixing up e and v?}
The result here is a vector $\bm{\alpha}'_i$, with one element for each child node $j \in \mathit{children}(i)$, and which sums to one.

On Yelp dateset, this variant of the \fullm model achieves 70.3\% accuracy (1.5\% absolute behind our \fullm model), giving empirical support to our theoretically-motivated design decision not to normalize attention.  
Of course, further architecture improvements may yet be possible.

\paragraph{Discussion.} Our findings in this work show the benefit of using discourse structure for text categorization. 
  Although discourse structure strongly improves the performance on most of corpora in our experiments, its benefit is limited particularly by two factors: (1) the state-of-the-art performance on RST discourse parsing; and (2) domain mismatch between the training corpus for a discourse parser and the domain where the discourse parser is used.
  For the first factor, discourse parsing is still an active research topic in NLP, and may yet improve.
  The second factor suggests exploring domain adaptation methods or even direct discourse annotation for genres of interest.

% ------------------------------
% Related work
\section{Related Work}
\label{sec:related}

Early work on text categorization often treated text as a bag of words \citep[e.g.,][]{joachims1998text,yang1997comparative}.
Representation learning, for example through  matrix decomposition \citep{deerwester1990indexing} or latent topic variables \citep{ramage2009labeled}, has been considered to avoid overfitting in the face of sparse data.

% Text categorization is esentially a classification task. 
% Once a text is represented as a numeric vector, the last step is to train a classifier to perform the categorization task. 
% Besides choosing a suitable classification model, from the perspective of NLP, how to represent a text is the central problem. 
% Early work on text categorization mostly uses bag-of-words representation or bag-of-features representation~\citep{joachims1998text}.
% Both of them represent a text with a collection of words or surface-form features (e.g., POS tags). 
% Due to the problem of overfitting with sparse representation, researchers prefer to use dense representation based on either latent semantic analysis~\citep{deerwester1990indexing} or topic models~\citep{ramage2009labeled}.

The assumption that all parts of a text should influence categorization equally persists even as more powerful representation learners are considered.
\citet{zhang2015character} treat a text as a sequence of characters, proposing to a deep convolutional neural network to build text representation. 
\citet{xiao2016efficient} extended that architecture by inserting a recurrent neural network layer between the convolutional layer and the classification layer. 

In contrast, our contributions follow \citet{ko2004improving}, who sought to weight the influence of different parts of an input text on the task.
Two works that sought to learn the importance of sentences in a document are \citet{yessenalina2010multi} and \citet{yang2016hierarchical}.  
The former used a latent variable for the informativeness of each sentence, and the latter  used a neural network to learn an attention function.  Neither used any linguistic bias, relying only on task supervision to discover the latent variable distribution or attention function.
Our work builds the neural network directly on a discourse dependency tree, favoring the most central EDUs over the others but giving the model the ability to overcome this bias.

Another way to use linguistic information was presented by \citet{yogatama2014linguistic}, who used a bag-of-words model.  The novelty in their approach was a data-driven regularization method that encouraged the model to collectively ignore groups of features found to coocur.  Most related to our work is their ``sentence regularizer,'' which encouraged the model to try to ignore training-set sentences that were not informative for the task.  Discourse structure was not considered.

\paragraph{Discourse for sentiment analysis.} 
Recently, discourse structure has been considered for sentiment analysis, which can be cast as a text categorization problem.
\citet{bhatia2015better} proposed two discourse-motivated models for sentiment polarity prediction.  One of the models is also based on discourse dependency trees, but using a hand-crafted weighting scheme. 
% \citet{hogenboom2015using} similarly used RST and compared six different hand-crafted weighting schemes. 
% They also evaluated on the P\&L corpus, and but considered only lexicon-based sentiment analysis, therefore their best performance is below 72\%.
% Table 1 shows that fully supervised methods give much stronger performance on this dataset, with accuracies more than 10\% higher.
% \nascomment{why can't we compare with them?  I looked at the paper and it's a total fishing expedition.  I couldn't figure out what dataset they used, so it may not make sense.  but we should explain somehow to reviewers why we aren't considering them for comparison.}
Our method's attention mechanism automates the weighting.
% \todo{Sentiment analysis with other discourse information}

% \paragraph{Dependency structure for composition}
% The first work on dependency structure for semantic composition is the Dependency-Tree Recursive Neural Network (DT-RNN) proposed in \citet{socher2014grounded}, where dependency structure is used to composite sentence representation via word embeddings.
% The immediate difference is that our models work on the text level.
% However, regardless the granularity in composition, the major distinctions are (1) the \fullm uses relation-specific instead of position-specific composition matrices; (2) no content-based attention weight is used in DT-RNN.
% \todo{More work on recursive NNs with dependencies.}

% \nascomment{are there other recursive NNs based on dependencies, or even other tree structures, that are worth mentioning?  readers might wonder why we didn't use something more like the phrase-structure recursive NNs in other Socher papers (and others' work), since that's closer to the original RST representation.  we should address that.}

% -------------------------------
% Conclusion
\section{Conclusion}
\label{sec:con}

We conclude that automatically-derived discourse structure can be
helpful to text categorization, and the benefit increases with the
accuracy of discourse parsing.  We did not see a benefit for categorizing legislative
bills,  a text genre whose discourse structure
diverges from that of news.  These findings
motivate further improvements to discourse parsing, especially for
new genres.

% ------------------------------
% Ack
\section*{Acknowledgments}
  We thank anonymous reviewers
 and  members of Noah's ARK for helpful feedback on this work. 
  We thank Dallas Card and Jesse Dodge for helping prepare the Media Frames Corpus and the Congressional bill corpus.
  This work was made possible by a University
  of Washington Innovation Award.

% *****************************************************************
\bibliographystyle{acl_natbib}
\bibliography{myref}

\clearpage
\appendix
\section{Supplementary Material: An example text from the  Bill corpus}
\label{append:example}

\begin{tabular}{p{0.95\textwidth}}
  \toprule
  4449 IH \\
  103d CONGRESS\\
  2d Session\\
  H. R. 4449\\
  To amend part A of title IV of the Social Security Act to enable States to construct, rehabilitate, purchase or rent permanent housing for homeless AFDC families, using funds that would otherwise be used to provide emergency assistance for such families.\\
  IN THE HOUSE OF REPRESENTATIVES\\
  MAY 18, 1994\\
  Mr. PETERSON of Minnesota (for himself, Mr. FLAKE, Mr. FRANK of Massachusetts, Mr. VENTO, and Mr. RANGEL) introduced the following bill; which was referred jointly to the Committees on Ways and Means and Banking, Finance and Urban Affairs\\
  A BILL\\
  To amend part A of title IV of the Social Security Act to enable States to construct, rehabilitate, purchase or rent permanent housing for homeless AFDC families, using funds that would otherwise be used to provide emergency assistance for such families.\\
  Be it enacted by the Senate and House of Representatives of the United States of America in Congress assembled,\\
  SECTION 1. SHORT TITLE.\\
  This Act may be cited as the `Permanent Housing for Homeless Families Act'.\\
  SEC. 2. EMERGENCY ASSISTANCE DEEMED TO INCLUDE CONSTRUCTION, REHABILITATION, PURCHASE, AND RENTAL OF PERMANENT HOUSING FOR HOMELESS AFDC FAMILIES.\\
  (a) IN GENERAL- Section 406 of the Social Security Act (42 U.S.C. 606) is amended by inserting after subsection (c) the following:\\
  `(d)(1) The term `emergency assistance to needy families with children' includes the qualified expenditures of an eligible State.\\
  `(2) As used in paragraph (1):\\
  `(A) The term `eligible State' means, with respect to a fiscal year, a State that meets the following requirements:\\
  `(i) The State plan approved under this part for the fiscal year includes provision for emergency assistance as described in subsection (e) or this subsection.\\
  `(ii) The State has provided assurances to the Secretary that the average amount that the State intends to expend per family for such emergency assistance for the fiscal year would not exceed such average amount for the immediately preceding fiscal year. The Secretary shall prescribe in regulations standards for determining the period over which capital expenditures incurred in the provision of such emergency assistance are to be amortized.\\
  $\dots$\\
  \bottomrule
\end{tabular}

\end{document}